\documentclass[runningheads]{llncs}
\usepackage{graphicx,amsmath,amssymb}
\usepackage{algorithm} 
\usepackage{algorithmic}  
\usepackage[algo2e]{algorithm2e}
\usepackage[numbers,sort&compress]{natbib}
\usepackage{hyperref}       
\hypersetup{colorlinks,linkcolor={blue},citecolor={blue},urlcolor={red}}  
\usepackage{url}            
\usepackage{booktabs}
\begin{document}
\title{Scalable Neural Architecture Search for 3D Medical Image Segmentation}
%
%
\author{Sungwoong Kim\inst{1}, Ildoo Kim\inst{1}, Sungbin Lim\inst{1}, Woonhyuk Baek\inst{1} \\ Chiheon Kim\inst{1}, Hyungjoo Cho\inst{2}, Boogeon Yoon\inst{1}, Taesup Kim\inst{3}}
\authorrunning{S. Kim et al.}

\institute{Kakao Brain, Pangyo, Seongnam, Gyeonggi, South Korea \\ 
\email{\{swkim, ildoo.kim, sungbin.lim, wbaek, chiheon.kim, eric.yoon\}@kakaobrain.com} \\
\and
Department of Transdisciplinary Studies, Seoul National University, South Korea \\
\email{joysquare@snu.ac.kr}\\
\and
MILA, Universit\'e de Montr\'eal, Canada\\
\email{taesup.kim@umontreal.ca}}

\maketitle              
\begin{abstract}
In this paper, a neural architecture search (NAS) framework is proposed for 3D medical image segmentation, to automatically optimize a neural architecture from a large design space. Our NAS framework searches the structure of each layer including neural connectivities and operation types in both of the encoder and decoder. Since optimizing over a large discrete architecture space is difficult due to high-resolution 3D medical images, a novel stochastic sampling algorithm based on a continuous relaxation is also proposed for scalable gradient based optimization. On the 3D medical image segmentation tasks with a benchmark dataset, an automatically designed architecture by the proposed NAS framework outperforms the human-designed 3D U-Net, and moreover this optimized architecture is well suited to be transferred for different tasks.

\end{abstract}

\section{Introduction}
\label{sec:intro}
Recently, deep neural networks have been extensively used for medical image segmentation tasks. However, in general, their performance relies on manual trial-and-error processes for making decisions on the network architecture, hyperparameters for training, and pre-/post-procedures. Due to being restricted to manual tuning, they would have limitations in performance improvement as well as fast transfer to related tasks. Currently, the same problem in the field of general deep learning has promoted the rapid development of automated machine learning (AutoML). Yet, in contrast to the recent studies on the use of advanced AutoML algorithms such as neural architecture search (NAS) \cite{liu2018darts,pham2018efficient,zoph2017learning} and neural optimizer search \cite{Wichrowska17} for general computer vision tasks, only few approaches using simple hyperparameter optimization have been proposed for medical imaging tasks \cite{Mortazi18,Naceur18}. 

\begin{figure}[t]
    \centering
    \includegraphics[scale=0.18]{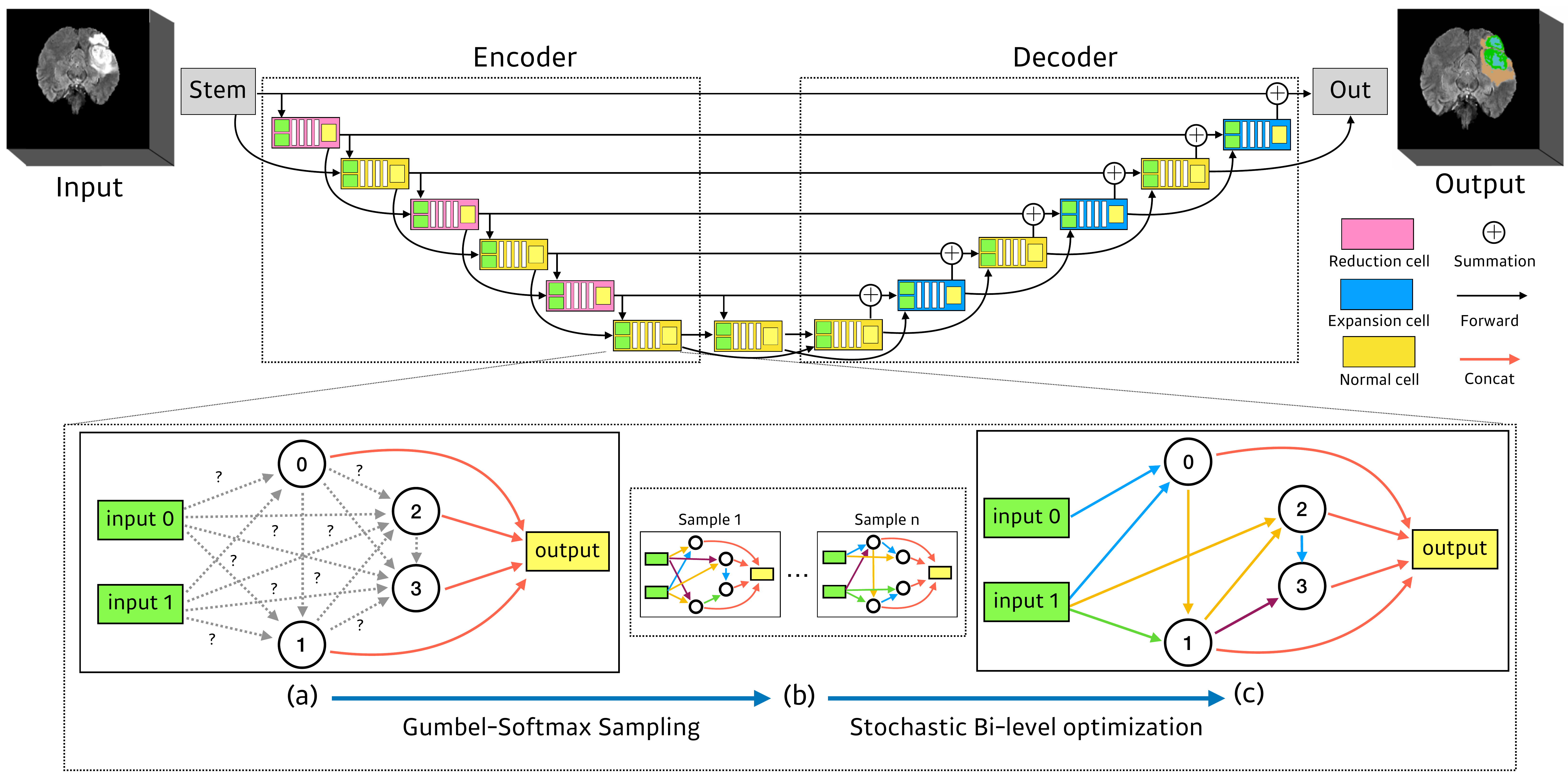}
    \caption{An overview of architecture search space for image segmentation tasks. Both encoder and decoder alternately stack normal cells and resizing (reduction, expansion) cells. The directed arrows between cells indicate the forward paths. Each cell is represented as a directed acyclic graph (DAG). (a) Initial candidate operations on the edges in DAG. (b) Gumbel-softmax operation sampling on each edge. (c) Induced final architecture from the obtained solution.}
    \label{fig1:encoder-decoder}
    \vspace{-0.5cm}
\end{figure}

In this paper, we propose a NAS framework for AutoML in designing neural networks especially for 3D medical image segmentation. 3D U-Net \cite{CicekALBR16} has been popularly used for segmenting high-resolution 3D medical images (see \cite{KayalibayJS17,Oktay18,Yu17}) since both semantic as well as spatial information can be efficiently exploited through skip connections between an encoder and a decoder. However, a convolutional block for each layer in the 3D U-Net has been manually designed with various convolutional filter types, pooling types, skip-connections, and non-linear activation functions. Instead of using the artificial block, we employ a NAS framework to obtain an automatically tuned structure of the block, which is called a cell, for each layer in the 3D U-Net where all cell structures and the corresponding neural operation parameters (e.g. kernel weights) are simultaneously learned in the end-to-end manner. For this, four types of cells, \textit{encoder-normal}, \textit{reduction}, \textit{decoder-normal}, and \textit{expansion} cell are defined to compose the encoder as well as the decoder for the learned U-Net architecture (see Figure \ref{fig1:encoder-decoder}). This is different from previous NAS approaches which merely use two types of cells (normal and reduction) for encoder-only networks \cite{liu2018darts,pham2018efficient}.

It is important to employ a sufficiently large search space in NAS in generating an improved network architecture on a target task. However, optimization over such a large space is difficult due to the extreme memory usage and the long run-time when dealing with high-resolution 3D images. Moreover, an exact bi-level formulation of NAS has to be solved on the mixed domain of (1) discrete variables regarding neural connections and operation types in each cell and (2) continuous operation parameters. This constraint restricts the use of a gradient method for architecture searching. To handle this problem, we propose a novel continuous approximation using Gumbel-softmax \cite{jang2016categorical} sampling on the discrete variables. This makes it possible to use a stochastic gradient descent (SGD) in bi-level optimization. This sampling procedure also enables to reduce the computational burden of taking the entire connectivities and operations into account within an outrageously large network originated from the continuous relaxation. Namely, the proposed differentiable NAS with stochastic sampling supports great scalability in terms of solvable large search space with reduced computational cost. To our best knowledge, this is the first work to exploit a complete NAS framework for automatically designing an architecture for the task of 3D medical image segmentation. 

Experimental results on the benchmark 3D medical image segmentation dataset show that in comparison to the human-designed 3D U-Net \cite{CicekALBR16}, the network obtained by the proposed scalable NAS leads to better performances. It is furthermore shown that the found architecture from a task having large amounts of labeled data can be transferred to build a network for different segmentation tasks with various medical modalities including MRI and CT that have small amounts of labeled data and achieves better generalization performances.

\section{Method}
\label{sec:method}

In this section, we first describe an architecture search space for 3D medical image segmentation. Then, we present a SGD-based bi-level optimization to simultaneously learn both of the architecture and the corresponding neural operation parameters.

\subsection{Search Space for 3D Medical Image Segmentation}
\label{subsec:search-space}

Following the idea of micro search space popularly used in the state-of-the-art NAS approaches \cite{liu2018darts,zoph2017learning,pham2018efficient}, U-Net-like networks, which is composed of encoder and decoder layers, are designed as repeated encoder and decoder cells (see Figure \ref{fig1:encoder-decoder}). Here, a cell $C$ is one of four cell-types - \emph{encoder-normal} ($C_{\mathsf{enc}}$), \emph{reduction} ($C_{\mathsf{red}}$), \emph{decoder-normal} ($C_{\mathsf{dec}}$), and \emph{expansion} ($C_{\mathsf{exp}}$) - and the normal cells and resizing cells are stacked alternately with skip connections between the cells in the encoder and the cells in the decoder. Note that an \textit{inter-cell}, a copy of $C_{\mathsf{enc}}$, is deployed between the encoder and decoder. Every cell takes two outputs of the last previous two cells as inputs\footnote{In the decoder, before used as one of inputs of the current cell, an output of the last previous cell is summed with an output of the encoder cell at the same level.} except the first reduction cell which takes an output of the predefined first convolutional block, called a stem cell, and then duplicates it as two inputs. The segmentation output is obtained from the predefined last convolutional block, referred to as an out cell.

The neural structure in each cell $C \in \mathcal{C} := \{ C_{\mathsf{enc}}, C_{\mathsf{red}}, C_{\mathsf{dec}}, C_{\mathsf{exp}}\}$ is represented as a directed acyclic graph (DAG). Let ${\mathcal G}=\mathcal{G}(C) = (\mathcal{V},\mathcal{E})$ be the DAG where each node $i\in\mathcal{V}$ corresponds to an intermediate feature vector ${\bf x}^{i}$ in cell $C$, and each directed edge $(i,j)\in\mathcal{E}$ stands for a connection between nodes $i$ and $j$ with a certain operation $o^{(i,j)}\in \mathcal{O}$ such that $\mathcal{O}$ denotes the set of all candidate operations, and $\mathbf{x}^{j} = \sum_{(i,j)\in\mathcal{E}}o^{(i,j)}(\mathbf{x}^{i})$. The output of a cell is a channel-wise concatenation of all the intermediate nodes. Therefore, the architecture search problem now amounts to find the best combination of all edge operations in the four cell-types. Basically, even the same type of cells can have different structures according to their layer levels. However, in this work, for simplicity, all cells that have a common type share a common structure regardless of layer levels. Note that a \emph{zero} operation is also one of the candidate operations to optimize the neural connectivities as well; \emph{zero} means a disconnection between two nodes.

\subsection{Stochastic Bi-Level Optimization with Operation Sampling}
\label{subsec:stochastic bi-level}

We first represent the selected edge operation using the one-hot vector ${\bf z}^{(i,j)} = \{z_{o}^{(i,j)}\mid o\in\mathcal{O}\}$ and operation vector $\mathbf{o}^{(i,j)}=\{o(\mathbf{x}^{i};\theta_{o}^{(i,j)}) | o\in\mathcal{O}\}$:
\begin{align}
\label{eq:op-selection}
    o^{(i,j)}({\bf x}^{i}) = \langle \mathbf{z}^{(i,j)},\mathbf{o}^{(i,j)} \rangle= \sum_{o\in\mathcal{O}} z_{o}^{(i,j)} o({\bf x}^{i}; {\bf \theta}_{o}^{(i,j)}),
\end{align}
where ${\bf \theta}_{o}^{(i,j)}$ denotes the parameter set of the operation $o$ on edge $(i,j)$. Then, finding the best cell architecture corresponds to solving the following bi-level optimization problem:
\begin{equation}
\label{eq:objective}
\begin{aligned}
    \min_{Z} \qquad&  \mathcal{L}_{\mathsf{val}}(\Theta^{*}(Z), Z) \\
    \text{s.t.} \qquad& \Theta^{*}(Z) = \underset{\Theta}{\text{argmin }}\mathcal{L}_{\mathsf{train}}(\Theta, Z),
\end{aligned}
\end{equation}
where $(\Theta, Z) = \{ (\theta^{(i,j)}, {\bf z}^{(i,j)}) \mid (i,j) \in \mathcal{E}(C), C \in \mathcal{C}\}$, and $\mathcal{L}_{\mathsf{val}}$ and $\mathcal{L}_{\mathsf{train}}$ are the validation loss and training loss, respectively. Bi-level pragramming \eqref{eq:objective}, is hard to solve since its search space is the mixed domain of continuous variables $\Theta$ and discrete variables $Z$. DARTS \cite{liu2018darts} try to circumvent this difficulty by relaxing $Z$ to a continuous logits and considering mixed operations in the edges. This allows to use SGD method to obtain an approximate solution and derive the final architecture from the relaxed variables by taking the operation with the highest strength on each edge. However, this method is infeasible since training the mixed operations in edges requires the extremely large memory usage and the long run-time for desired high-resolution 3D image segmentation tasks.

To overcome the aforementioned problems, we propose a modified optimization, called \emph{stochastic bi-level optimization}, by first treating $Z$ as random discrete variables and then replacing \eqref{eq:objective} as
\begin{equation}
\label{eq:stochastic-objective}
\begin{aligned}
    \min_{\alpha} \qquad&  \mathbb{E}_{Z\sim P_{\alpha}}[\mathcal{L}_{\mathsf{val}}(\Theta^{*}(Z), Z)] \\
    \text{s.t.} \qquad&  \Theta^{*}(Z) = \underset{\Theta}{\text{argmin }} \mathcal{L}_{\mathsf{train}}(\Theta, Z),
\end{aligned}
\end{equation}
where $P_\alpha$ is the discrete distribution on $Z$, parameterized by continuous variable $\alpha$. Since it is intractable to exactly compute $\nabla_{\alpha} \mathbb{E}_{Z\sim P_{\alpha}}[\mathcal{L}_{\mathsf{val}}(\Theta^{*}(Z), Z)]$, we estimated it through continuous relaxation using the Gumbel-softmax reparametrization technique \cite{jang2016categorical} as
\begin{equation}
\label{eq:gumbel-softmax-reparametrization}
\begin{aligned}
\!\!\!\nabla_{\alpha} \mathbb{E}_{Z\sim P_{\alpha}}[\mathcal{L}_{\mathsf{val}}(\Theta^{*}(Z), Z)] \approx \mathbb{E}_{\epsilon \sim \mathsf{Gumbel}(0,1)} [\nabla_{\alpha}\mathcal{L}_{\mathsf{val}}(\Theta^{*}(\bar{Z}(\alpha,\epsilon; \tau)), \bar{Z}(\alpha,\epsilon; \tau)],\!\!
\end{aligned}
\end{equation}
where continuously relaxed variables $\bar{Z}(\alpha,\epsilon; \tau) = \mathsf{softmax}((\alpha + \epsilon)/\tau)$, $\tau$ denotes the temperature, and $\epsilon$ is random variable drawn from the Gumbel distribution. Here, the expectation in \eqref{eq:gumbel-softmax-reparametrization} is approximated with $\epsilon$-sampling. It is noted that as $\tau \rightarrow 0$, the distribution of $\bar{Z}$ is identical to $P_\alpha$, which means that by annealing $\tau$ we can enforce $\bar{Z}$ to be one-hot discrete variables $Z$ during training; the relaxed architecture is forced to be converged to the final architecture.

When alternatively updating $\Theta$ and $\alpha$ by respective gradient descents, we again replace $\bar{Z}$ with $\hat{Z}$ by sampling two operations on each edge from the Gumbel-softmax with rescaling of the corresponding two operation weights to be summed to one, as shown in Algorithm \ref{alg:gumbel-sample}. Note that due to $\tau$-annealing, the number of sampled operations on each edge is naturally reduced from two to one during training. This stochastic operation sampling supports improved scalability in terms of solvable large search space with small computational cost.

\begin{algorithm}[t]
\SetKwInOut{Input}{Input}

\Input{Logits $\alpha = (\alpha^{(i,j)}_{o}: (i,j) \in \mathcal{E}\text{ and } o \in \mathcal{O})$, temperature $\tau$}
\For{$(i,j) \in \mathcal{E}$}{
    $\bar{\bf z}^{(i,j)} \leftarrow \mathsf{softmax}((\alpha^{(i,j)} + \epsilon^{(i,j)})/\tau)$, $\epsilon_{o}^{(i,j)} \sim \mathsf{Gumbel}(0,1), o \in \mathcal{O}$

    \ForEach{pair $\{o_1, o_2\}$ in $\mathcal{O}$}{
    $q^{\{o_1, o_2\}} \leftarrow (\bar{z}_{o_1}^{(i,j)} + \bar{z}_{o_2}^{(i,j)})/(|\mathcal{O}|-1)$\;
    }
    Sample $\{o_1, o_2\}$ with $q^{\{o_1, o_2\}}$

    $\left(\hat{z}^{(i,j)}_{o_1}, \hat{z}^{(i,j)}_{o_2}\right)
    \leftarrow
    \left(\frac{\bar{z}^{(i,j)}_{o_1}}{\bar{z}^{(i,j)}_{o_1}+\bar{z}^{(i,j)}_{o_2}} \frac{\bar{z}^{(i,j)}_{o_2}}{\bar{z}^{(i,j)}_{o_1}+\bar{z}^{(i,j)}_{o_2}}\right)$, $\hat{z}^{(i,j)}_{o} \leftarrow 0, ~~ o \notin \{o_1, o_2\}$\;
}
\Return $\hat{Z}$\;
\caption{Gumbel Softmax Sampling}
\label{alg:gumbel-sample}
\end{algorithm}

\section{Experiments}
\label{sec:exp}

\subsubsection*{Dataset and Evaluation}
The proposed scalable NAS (SCNAS) was evaluated on the three 3D segmentation tasks, (1) brain tumor (MRI, 484 labeled images, 3 classes), (2) heart (MRI, 20 labeled images, 1 class), and (3) lung (CT, 64 labeled images, 1 class), from the Medical Segmentation Decathlon challenge (MSD, \url{http://medicaldecathlon.com}) where each task has different input modalities and sizes as well as different foreground classes, which is therefore suitable for evaluating the generalizability and transferability of the SCNAS. Since the ground-truth labels for test images are not provided in the MSD dataset, the evaluation was conducted by 5-fold cross-validation (CV) on the training images with the average dice similarity coefficient as the metric. Here, the authors in \cite{Isensee18} provided their splitting for this 5-fold CV, and we used it. For SCNAS, the training set after the validation split was split again into two sets with a ratio of 4:1 for respectively optimizing the operation parameters and the architecture parameters. 

\subsubsection*{Implementation Details}
The performances of the SCNAS are compared to those obtained by our baseline 3D U-ResNet \cite{KayalibayJS17,Yu17}, which makes use of residual blocks, multiple segmentation maps \cite{KayalibayJS17}, and attention gates \cite{Oktay18}, as well as those from the 3D nnU-Net \cite{Isensee18} that can be considered as the best performed single model from the perspective of challenge results. In both of the 3D U-ResNet and the SCNAS, patch-based training and inference were carried out such that each image was randomly cropped to the region of nonzero values with the predefined resolution during training, while in testing, the prediction results were obtained by combining patch-based inference results with 50 percent overlap. The input patch size was basically set to $128\times128\times128$ and modified for each task taking median shapes and memory constraints into account just like that used for the 3D nnU-Net in \cite{Isensee18}. Since even the same task provides 3D images with heterogeneous voxel spacings, the input images were first resampled to have an equal voxel spacing of 0.7mm $\times$ 0.7mm $\times$ 0.7mm, and then $z$-normalization was separately applied to each input channel. Following \cite{Isensee18}, we also utilized the data augmentation techniques at both training and testing time with the the same kinds and parameters that used in \cite{Isensee18}. However, unlike \cite{Isensee18}, network-cascade, prediction-ensembling from different architectures, and the removal of small connected components were not adopted in this evaluation to solely examine the effects by the use of NAS in designing the network architecture.

The set of operations $\mathcal{O}$ on each edge in the SCNAS consists of the following eight operations: $3\times 3\times 3$ convolutions, depthwise separable dilated $3\times 3\times 3$ convolutions with rate 2, 3 and 4, $3\times 3 \times 3$ max and average 3D pooling, identity (skip connection), and zero. Here, we used the LeakyReLU-Conv-InstanceNorm for convolutional operations. As shown in Figure \ref{fig1:encoder-decoder}, the whole network in the SCNAS is composed of 12 automatically designed cells, each of which has 4 nodes. This number of stacked cells is consistent with that of the 3D U-ResNet in terms of respective three times of downsampling and upsampling by a factor of 2. Here, all operations in the reduction cell in the SCNAS are of stride two while the expansion cells perform pre-upsampling for the inputs of the cell. Similar to the 3D U-ResNet, the reduction and expansion cells in the SCNAS respectively double and halve the number of output channels of given inputs.

It it noted that the SCNAS first optimized all of cell architectures using 48 output channels of the stem cell in order to fit a batch size of 1 into a single GPU. Then, a larger network was constructed by increasing the number of stem channels to 68 with found cell-topologies and was retrained from scratch. Here, 68 channels makes the computational complexity for inference of a found network by SCNAS to be similar to the baseline 3D U-ResNet, which has 32 output channels in the first convolutional blolck, in terms of FLOPs: 419.59 GFLOPs (3D U-ResNet) vs. 424.76 GFLOPs (SCNAS) on the brain tumor task.

The SCNAS was trained for 200 epochs with a batch size of 4, which took one day on 4 V100 GPUs. In this SCNAS training, the ADAM optimizer were used where the initial learning rates / beta parameters were as set to be $0.025$ / $(0.1, 0.001)$ for training operation parameters $\Theta$ and $0.003$ / $(0.5, 0.999)$ for training architecture parameters $\alpha$. If a plateau for 20 epochs on the training loss was detected, the learning rate was reduced by a factor of 10. When retraining the SCNAS models as well as training the 3D U-ResNet models, an initial learning rate of $0.0003$ and beta parameters of $(0.9, 0.999)$ for the ADAM optimizer were used with a batch size of 8, where the learning rate was reduced by a factor of 5 if a training loss was not reduced for 30 epochs, and the iteration was terminated either if it lasted for 500 epochs or if the learning rate was smaller than $10^{-7}$. The loss function for both 3D U-ResNet and SCNAS is the Jaccard distance \cite{KayalibayJS17}.

\begin{table*}[t!]  \center 
\caption{Average dice similarity coefficients (\%) on three tasks of MSD. \cite{Isensee18} obtained their 3D nnU-Net results by model selection based on the validation loss.}
\label{table1:dice}
\begin{tabular}{c | c c c c | c | c }
\toprule
& \multicolumn{4}{c|}{Brain Tumor (MRI)} & \multicolumn{1}{c|}{Heart (MRI)} & \multicolumn{1}{c}{Lung (CT)} \tabularnewline
\hline
\scriptsize{Label} & \scriptsize{Edema} & \scriptsize{Non-Enhancing} & \scriptsize{Enhancing} & \scriptsize{Average} & \scriptsize{Left Atrium} & \scriptsize{Tumor} \tabularnewline
\hline
3D nnU-Net \cite{Isensee18} & 80.71 & 62.22 & 79.07 & 74.00 & 92.45 & 55.87 \tabularnewline
3D U-ResNet & 70.74 & 56.69 & 73.23 & 66.89 & 91.48 & 63.28 \tabularnewline
\midrule
SCNAS & 80.41 & 59.85 & 78.50 & 72.92 & 91.29 & 64.82 \tabularnewline
SCNAS(transfer) & - & - & - & - & 91.91 & 68.62 \tabularnewline 
\bottomrule
\end{tabular}
\end{table*}

\subsubsection*{Results}
\begin{figure}[t]
    \centering
    \includegraphics[scale=0.18]{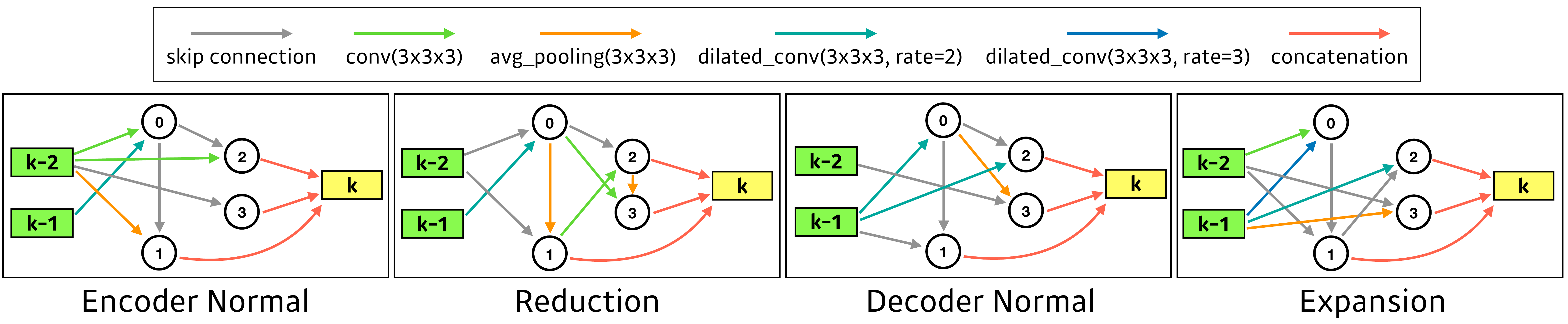}
    \caption{The found cell architectures by SCNAS on the brain tumor task (first CV fold).}
    \label{fig2:genotype}
\end{figure}

\begin{figure}[t]
    \centering
    \includegraphics[scale=0.18]{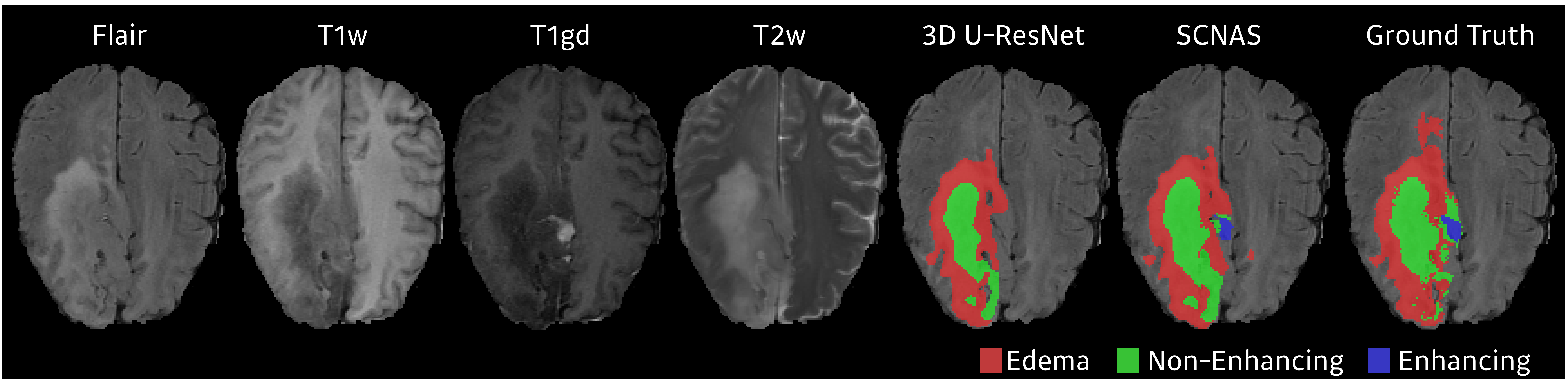}
    \caption{Segmentation results on the brain tumor task from the 3D U-ResNet and the SCNAS. The first four images from left show the MRI sequences used as input channels. The last image shows the ground truth.}
    \label{fig3:prediction}
\end{figure}

Table \ref{table1:dice} shows that the SCNAS produced better architectures than the (human-designed) 3D U-ResNet in terms of the overall performances. Especially, the performances of SCNAS are comparable or even better than those of the 3D nnU-Net \cite{Isensee18}. Here, it should be noted that the 3D nnU-Net performed model selection based on the validation loss during their 5-fold CV while ours did not take any validation result into account during training. On the heart and lung segmentation tasks, which have only 20 and 64 labeled images, respectively, the 3D U-ResNet as well as the SCNAS can be prone to overfitting on the training set. Therefore, we transferred the found architecture by SCNAS from the first CV fold of the brain tumor task having 484 labeled images to these tasks. For this, we modified the stem cell architecture to match the number of input channels according to each task, and the operation parameters in the transferred architecture were retrained from scratch on each task. As a result, the transferred architecture from the brain tumor task achieved better generalization performances in comparison to their own NAS results. Figure \ref{fig2:genotype} shows the optimized cell architectures by SCNAS on the brain tumor task. We conjecture that the selected dilated convolutions are helpful to reflect a more global context for improving segmentation results. Example input images and the corresponding segmentation outputs from the brain tumor task are presented in Figure \ref{fig3:prediction}, which shows better segmentation results by SCNAS compared to 3D U-ResNet.

\vspace{-0.2cm}
\section{Conclusion}
\label{sec:conclusion}

In this work, a complete NAS framework for automatically designing an architecture is proposed and demonstrated on the benchmark dataset of 3D medical image segmentation tasks. In the proposed framework, NAS is formulated as finding the optimal structure of four types of cells composing an encoder as well as a decoder. We introduce a novel stochastic sampling algorithm which results in significant improvement in terms of the scalability suitable for handling high-resolution 3D medical images. Empirical evaluation demonstrates that the automatically optimized network via the proposed NAS outperforms the manually designed 3D U-Net, and the learned architecture is successfully transferred to different segmentation tasks.

\bibliographystyle{splncs04}

\end{document}